\documentclass[letterpaper, 10 pt, conference]{ieeeconf}

\IEEEoverridecommandlockouts                             
\usepackage[english]{babel}
\let\labelindent\relax
\usepackage{enumitem}
\usepackage[colorlinks=true, allcolors=blue]{hyperref}
\usepackage{amsmath}
\usepackage[english]{babel}
\usepackage{amssymb}
\usepackage{tablefootnote}
\usepackage{mathtools}
\usepackage{svg}
\usepackage{algorithm} 
\usepackage{algpseudocode} 
\usepackage{makecell, multirow}
\usepackage{graphicx}
\usepackage{caption}
\usepackage{subcaption}
\usepackage{hyperref}
\usepackage{comment}
\usepackage{amsfonts} 
\usepackage{cleveref}
\usepackage{bm}
\usepackage{multirow}
\usepackage{algorithm}
\usepackage{booktabs}
\usepackage{color}
\usepackage{verbatim}
\usepackage{adjustbox}
\usepackage{tikz}
\usetikzlibrary{calc}

\graphicspath{ {imgs/} }

\newtheorem{theorem}{Theorem}
\newcommand{\rgpSet}{\mathcal{G}_{\text{S}}}
\newcommand{\trajSet}{\mathcal{T}_{\text{S}}}
\newcommand{\safeDistanceSeq}{\mathcal{D}_{\text{s}}}
\newcommand{\uncertainAnalysis}{\mathcal{U}_{\text{a}}}
\newcommand{\traj}{\mathcal{T}}
\newcommand{\rIns}{r_{\emph{ins}}}
\newcommand{\gap}{G}
\newcommand{\funclabel}[1]{%
  \@bsphack
  \protected@write\@auxout{}{%
    \string\newlabel{#1}{{\jayden@currentfunction}{\thepage}}%
  }%
  \@esphack
}

\author{{Hongyi Chen$^{1}$, Shiyu Feng$^{2}$, Ye Zhao$^{2}$, Changliu Liu$^{3}$, Patricio A. Vela$^{1}$}
\thanks{$^{1}$ H. Chen and P.A. Vela are with the School of Electrical and Computer
        Engineering, Georgia Institute of Technology, Atlanta, GA 30308, USA.
        {\tt\small \{hchen657, pvela\}@gatech.edu}}%
\thanks{$^{2}$ S. Feng and Y. Zhao are with the School of Mechanical Engineering, Georgia Institute of Technology, Atlanta, GA 30308, USA.
        {\tt\small shiyufeng@gatech.edu, ye.zhao@me.gatech.edu}}%
\thanks{$^{3}$ C. Liu is with the Robotics Institute, Carnegie Mellon University, 5000 Forbes Ave, Pittsburgh, PA 15213, USA   {\tt\small cliu6@andrew.cmu.edu}}
}%
\title{Safe Hierarchical Navigation in Crowded Dynamic Uncertain Environments}

\begin{document}

\maketitle

\begin{abstract}
This paper describes a hierarchical solution consisting of a multi-phase planner and a low-level safe controller to jointly solve the safe navigation problem in crowded, dynamic, and uncertain environments. The planner employs dynamic gap analysis and trajectory optimization to achieve collision avoidance with respect to the predicted trajectories of dynamic agents within the sensing and planning horizon and with robustness to agent uncertainty. To address uncertainty over the planning horizon and real-time safety, a fast reactive safe set algorithm (SSA) is adopted, which monitors and modifies the unsafe control during trajectory tracking. Compared to other existing methods, our approach offers theoretical guarantees of safety and achieves collision-free navigation with higher probability in uncertain environments, as demonstrated in scenarios with 20 and 50 dynamic agents. Project website: \href{ https://hychen-naza.github.io/projects/HDAGap/}{ https://hychen-naza.github.io/projects/HDAGap/}.

\end{abstract}

\section{Introduction}
Deploying mobile robots ubiquitously requires that they safely and reliably accomplish navigation tasks in crowded, dynamic, and uncertain real world settings. These settings are challenging since the robot system is expected to plan online, handle the uncertainty, and establish safe actions to avoid multiple moving agents \cite{navi_problem_survey}.  Current progress towards this goal draws from hierarchical navigation, control theory, deep learning, and optimization. This paper leverages progress on these fronts to establish an online, hierarchical approach to trajectory synthesis in crowded, dynamic environments. 

Hierarchical navigation systems coordinate modules operating at different temporal and spatial scales \cite{Hierechical_Navigation_survey,fast_heuristic_planning,three-layer-navigation,PGap} to exploit the advantages of the chosen approaches while offsetting their limitations. Gap-based planners detect passable free-space in local environments while relying on an approximate global path planner.  However, current methods, like potential gap (PGap), are designed for static environments without dynamic agents \cite{PGap,closest_gap,gap_avoidance,egoTEB}. Optimization-based planning methods generate collision-free optimal trajectories as long as the objective function and constraints are well defined. The challenge lies designing optimization problems with real time computation properties \cite{CFS}. Reactive algorithms, like the potential field method (PFM) \cite{PFM}, control barrier functions (CBF) \cite{CBF}, and the safe set algorithm (SSA) \cite{SSA}, only consider one-step safe control calculation and can get stuck in local minima \cite{RL_SSA}. Learning-based planning and navigation in crowded environments lack safety guarantees, even if there is an extensive training phase \cite{RL_Nav,RL_NAV_2,RL_NAV_3}. 

We design a hierarchical navigation solution consisting of a high-level planner for long-term safety and robustness, and a low-level controller for online short-term safety guarantee. The planner layer itself is hierarchical and consists of three components. First, the proposed dynamic agents gap (DAGap) method handles spatio-temporally evolving gaps and synthesizes trajectories for detected gaps.  While the DAGap-generated trajectory considers specific pairwise agent groupings, the top two candidates warm start the convex feasible set (CFS) optimizer, which enforces hard safety constraints for all sensed agents while minimizing a trajectory optimizing objective function. To improve robustness, a proposed uncertainty analysis module estimates high-confidence bounds on the prediction errors of agents' positions for influencing a safety length parameter. At the controller layer, adopting the fast reactive safe set algorithm (SSA) monitors and modifies online any unsafe actions to reduce collisions caused by computation delay or trajectory tracking errors. The key contributions are summarized below:  
\begin{itemize}[leftmargin=*]
  \item Dynamic agent gap analysis for simplifying the candidate,
    spatio-temporally evolving solution space and synthesizing
    candidate trajectories.
  \item Hierarchical use of DAGap multi-trajectory synthesis followed by 
    CFS trajectory optimization for scaling the agents under consideration.
  \item High-confidence error bound estimation for use by the safety
    components of DAGap planning and CFS optimization, with provably high
    probability safety in uncertain environments.
  \item Analysis and benchmarking of the proposed solution relative to
    two hierarchical navigation methods ARENA \cite{ARENA} and DRRT-ProbLP
    \cite{DRRT-ProbLP}, and empirically shown to be safer.
\end{itemize}

\section{Related Work}

\subsection{Path Planning in Dynamic Environments}
Robotic path planning in static environments is a thoroughly studied problem that can typically be solved very efficiently. However, planning in the dynamic environment, especially in crowded dynamic environment, is still challenging because time is added as an additional dimension to the search-space and requires real-time replanning to deal with unprecedented situations in the future. To overcome the online computation challenges, dynamic A* and other incremental variants of classical planning are proposed which can correct previous solutions when updated information is received, so that the ego vehicle can safely interact with several dynamic agents \cite{D*}\cite{multi-resolution}\cite{MP-RRT}. However, they assume the agents to be static and rely on repeatedly replanning to generate collision-free paths, which may cause suboptimality. Restricting the search space by filtering out unsafe subspace is also very common. For example, SIPP eliminates collision intervals and searches in contiguous safe intervals \cite{SIPP} and ICS filters out the inevitable collision states when searching the waypoints \cite{ICS}. Besides, sampling-based planning is a classic concept and various strategies are proposed to handle moving agents \cite{BRANICKY2005271}, including sampling the control from robot’s state $\times$ time space \cite{hsu2002randomized} and sampling the robot movement based on the distribution of target goal and agents \cite{DRRT-ProbLP}. Optimization-based algorithms are used to compute smooth and collision-free paths in dynamic environments by setting proper cost functions and constraints \cite{CFS}\cite{TO_1}\cite{TO_2}\cite{7539623}. The downside is that the optimization problem is highly nonconvex which comes from the highly nonlinear inequality constraints, and is computationally expensive.

\subsection{Reactive Collision Avoidance}
Reactive approaches are extensively studied to compute an immediate action that would avoid collisions with obstacles. Energy-based reactive methods, include CBF \cite{CBF_Navi}, SSA \cite{SSA} and sliding mode algorithm \cite{SMA}, usually design a scalar energy function to achieve set-invariant control. The underlying assumption is that the dynamics of the system should be known. With the dynamic information, they can correctly and quickly solve an optimization problem to drive the energy function in the negative direction whenever the system state is outside of the safe set. Different from energy-based methods, gradient-based reactive methods like potential field method and its variants don't need the knowledge about system dynamics and offer simple computations \cite{PGap}\cite{APF_2}\cite{APF_3}. However, they lack of consideration for robot kinematics and dynamics and do not have sound safety guarantees. Optimal reciprocal collision avoidance \cite{ORCA} and reciprocal velocity obstacles \cite{RVO} derive the collision-free motion based on the definition of velocity obstacles. But they assume all agents' movements follow certain policies while this assumption is not always valid in the real world. Besides, reinforcement learning methods are becoming popular for safe navigation in crowded environments, however, purely learning based methods lack the safety guarantee even after long time training \cite{RL_Nav}\cite{RL_NAV_2}\cite{RL_NAV_3}. Recent safe learning algorithms use CBF or SSA as action monitor to keep modifying actions generated from the policy and achieve a low collision rate for safety-critical tasks \cite{RL_SSA}\cite{CBF_RL}. When these reactive methods are tested in challenging environments which require long-term planning, however, they may stuck in local minimal or lead to oscillations and cause collision because of their myopic property. To offset these limitations, some studies build hierarchical path planning consisting a low-level collision avoidance controller and a high-level planner \cite{three-layer-navigation}\cite{ARENA}.

\subsection{Gap-based Navigation}
Local planners using the representations of perception space can gain computational advantages by minimally processing the sensor data and recasting local navigation as an egocentric decision process \cite{Egocentric}\cite{egoTEB}. Following this idea, gap-based approaches aim at detecting passable free-space, which is defined as a set of ``gaps" comprised of beginning and ending points, from 1D laser scan measurements. Because of the detected collision-free regions, gap-based methods are compatible with other hierarchical navigation strategies to improve the safety of the synthesized trajectories \cite{closest_gap}\cite{gap_avoidance}. For example, egoTEB combines the representation of gap regions with the trajectory optimization method timed-elastic-bands (TEB) \cite{TEB}, which produces and optimizes multiple trajectories with distinct topologies \cite{egoTEB}. As a soft-constraint optimization approach, however, egoTEB cannot guarantee that optimized trajectory will fully satisfy all constraints and the poses of a trajectory may jump over an obstacle. Besides, the potential gap approach considers the integration of gap-based navigation with artificial potential field (APF) methods to derive a local planning module that has provable collision-free properties \cite{PGap}. However, all previous gap-based navigation methods are designed for static environments without dynamic agents. The intent of this study is to explore more deeply the gap representations in environment with crowded dynamic agents. 

\section{Methodology}
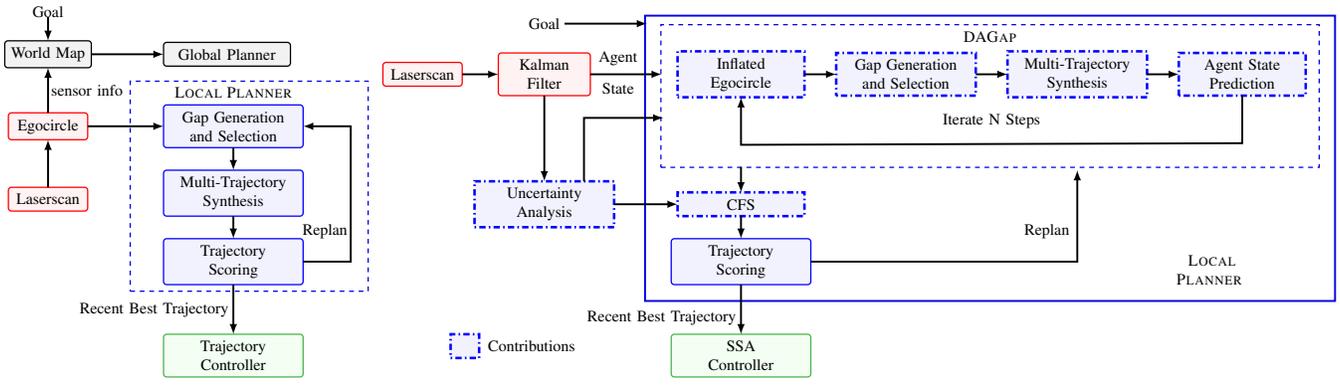
\begin{figure*}[htbp]
\vspace{0.5em}

  \hspace*{-0.4in}
  \scalebox{0.6}{\tikzstyle{block} = [draw, rectangle, text centered, thick,rounded corners=2pt,
                     minimum height=1.5em, minimum width=5em, inner sep=4pt]
\tikzstyle{typical} = [fill=white!95!black]
\tikzstyle{reddish} = [draw=red,fill=white!95!red]
\tikzstyle{blueish} = [draw=blue,fill=white!95!blue]
\tikzstyle{greenish} = [draw=green!40!gray,fill=white!95!green]
\tikzstyle{longblock} = [draw,rectangle,text centered,thick,rounded corners=2pt,
                     minimum height=1.5em, minimum width=8em, inner sep=4pt]
\tikzstyle{largeBlock} = [draw, rectangle, very thick,
                     minimum height=19.3em, minimum width=25em, inner sep=4pt]
\tikzstyle{smallBlock} = [draw, rectangle, text centered, thick, dashed,
                     minimum height=13.25em, minimum width=15em, inner sep=4pt]
\tikzstyle{dashedBlock} = [draw, dashed, rectangle,
                     minimum height=2em, minimum width=4em, inner sep=4pt]
\tikzstyle{dottedBlock} = [draw, rectangle, text centered, ultra thick,
					 minimum height=1.5em, 
					 minimum width=8em, inner sep=4pt,
					 dash pattern=on 1pt off 2pt on 4pt off 2pt] 
\tikzstyle{newtip} = [->, very thick]
\tikzstyle{bidir} = [<->, very thick]
\tikzstyle{newtip_dashed} = [->, very thick, dashed]
\begin{tikzpicture}[auto, inner sep=0pt, outer sep=0pt, >=latex]

  \node[block, reddish] (laserscan) {Laserscan};

    
  \node[block,reddish,anchor=south] (egocircle) 
    at ($(laserscan.north) + (0, 3em)$) 
    {\centering Egocircle};
  \node[block, typical, anchor=south] (worldmap)  
    at ($(egocircle.north)+(0, 1)$)
    {\centering World Map};
    
  \node[anchor=south] (goal) at ($(worldmap.north)+(0,1.5em)$) {Goal};

  \node[longblock, typical, anchor=west] (global) 
    at ($(worldmap.east) + (1.6, 0)$)  
    {\centering Global Planner};

  \node[longblock, blueish, anchor=west, text width=8em] 
    (gap) at (egocircle-|global.west)
    {\centering Gap Generation \\ and Selection};

  \node[longblock,blueish,anchor=north, text width=8em] (traj) 
    at ($(gap.south) - (0, 0.5)$) 
    {\centering Multi-Trajectory \\ Synthesis};
  
  \node[longblock,blueish,anchor=north, text width=8em] (score) 
    at ($(traj.south) - (0, 0.5)$) {\centering Trajectory \\ Scoring};
    
  \node[longblock,greenish,anchor=north, text width=8em] (controller) 
    at ($(score.south) - (0, 1.1)$) {\centering Trajectory \\ Controller};
    
  \node[smallBlock, draw=blue!90!black,anchor=north] (local) 
    at ($(gap)!0.09!(gap) + (1em, 1)$){};
  \node[anchor=north,xshift=-1em,yshift=-4pt] (localtext) at (local.north) 
      {\sc Local Planner};


  \draw[newtip] (laserscan.north) -- (egocircle.south);
  \draw[newtip] (egocircle.north) -- node[midway,right,xshift=2pt]{sensor info} 
    (worldmap.south);

  \draw[newtip] (worldmap.east) -- (global.west);
  \draw[newtip] (egocircle.east) -- (gap.west);

  \draw[newtip] (gap.south) -- (traj.north);
  \draw[newtip] (traj.south) -- (score.north);

  \draw[newtip] (score.south) -- node [midway,left,xshift=-3pt]{Recent Best Trajectory}(controller.north);
  
  \draw[newtip,<-] (gap.east) -- ($(gap.east)+(30pt,0pt)$) |-
  node[midway,left,anchor=east,xshift=-2pt,yshift=19pt]{Replan}(score.east);

  \draw[newtip] ($(goal.south)+(0pt,0pt)$) -- (worldmap.north);
%
%

  
\end{tikzpicture}}
  \hspace*{-0.4in}
  \scalebox{0.6}{\tikzstyle{block} = [draw, rectangle, text centered, thick,rounded corners=2pt,
                     minimum height=1.5em, minimum width=5em, inner sep=4pt]
\tikzstyle{typical} = [fill=white!95!black]
\tikzstyle{reddish} = [draw=red,fill=white!95!red]
\tikzstyle{blueish} = [draw=blue,fill=white!95!blue]
\tikzstyle{greenish} = [draw=green!40!gray,fill=white!95!green]
\tikzstyle{longblock} = [draw,rectangle,text centered,thick,rounded corners=2pt,
                     minimum height=1.5em, minimum width=8em, inner sep=4pt]
\tikzstyle{largeBlock} = [draw, rectangle, very thick,
                     minimum height=18em, minimum width=43.5em, inner sep=4pt]
\tikzstyle{smallBlock} = [draw, rectangle, text centered, thick, dashed,
                     minimum height=9em, minimum width=41.5em, inner sep=4pt]
\tikzstyle{dashedBlock} = [draw, dashed, rectangle,
                     minimum height=2em, minimum width=4em, inner sep=4pt]
\tikzstyle{dottedBlock} = [draw, rectangle, text centered, ultra thick,
					 minimum height=1.5em, 
					 minimum width=8em, inner sep=4pt,
					 dash pattern=on 1pt off 2pt on 4pt off 2pt] 
\tikzstyle{newtip} = [->, very thick]
\tikzstyle{bidir} = [<->, very thick]
\tikzstyle{newtip_dashed} = [->, very thick, dashed]
\begin{tikzpicture}[auto, inner sep=0pt, outer sep=0pt, >=latex]

  \node[block, reddish] (laserscan) {Laserscan};

    
  \node[block,reddish,anchor=west, text width=5em] (kalman) 
    at ($(laserscan.east) + (0.80, 0)$) 
    {\centering Kalman \\ Filter};
    
  \node[anchor=south] (goal) at ($(kalman.north)+(0,1.5em)$) {Goal};

  
  \node[dottedBlock, blueish, anchor=west, text width=5em] 
    (egocircle) at ($(kalman.east)+(5.5em,0)$)
    {\centering Inflated \\ Egocircle};
    
  \node[dottedBlock, blueish, anchor=west, text width=8em] 
    (gap) at ($(egocircle.east) + (2em, 0)$)
    {\centering Gap Generation \\ and Selection};

  \node[dottedBlock,blueish,anchor=west, text width=8em] (traj) 
    at ($(gap.east) + (2em, 0)$) 
    {\centering Multi-Trajectory \\ Synthesis};
    
  \node[dottedBlock,blueish,anchor=west, text width=6em] (prop) 
    at ($(traj.east) + (2em, 0)$) 
    {\centering Agent State \\ Prediction};
  
  \node[dottedBlock,blueish,anchor=north] (cfs) 
    at ($(egocircle.south) - (0, 6em)$) {\centering CFS};
    
  \node[dottedBlock,blueish,anchor=center, text width=8em] (uncert) 
    at (kalman.south|-cfs.west) {\centering Uncertainty \\ Analysis};
  
  \node[longblock,blueish,anchor=north, text width=8em] (score) 
    at ($(cfs.south) - (0, 0.5)$) {\centering Trajectory \\ Scoring};
    
  \node[longblock,greenish,anchor=north, text width=8em] (controller) 
    at ($(score.south) - (0, 1.1)$) {\centering SSA \\ Controller};
  
  \node[smallBlock, draw=blue!90!black,anchor=north] (dgap) 
    at ($(egocircle)!0.09!(egocircle) + (15.7em, 1.1)$){};
  \node[anchor=north,xshift=-0em,yshift=-4pt] (dgaptext) at (dgap.north) 
      {\sc DAGap};

  \node[largeBlock, draw=blue!90!black,anchor=north] (local) 
    at ($(egocircle)!0.09!(egocircle) + (15.7em, 1.3)$){};
  \node[anchor=south,xshift=14em,yshift=1em, text width=4.5em] (localtext) at (local.south) 
      {\centering {\sc Local \\ Planner}};


  \draw[newtip] (laserscan.east) -- (kalman.west);

  \draw[newtip] (kalman.east) -- ($(kalman.east) + (4.5em, 0)$)
  node[midway,align=center,anchor=center,xshift=-0.5em,yshift=0.1em, text width=3em]{\centering Agent \\[0.8em] State};

  \draw[newtip] (egocircle.east) -- (gap.west);
  \draw[newtip] (gap.east) -- (traj.west);
  \draw[newtip] (traj.east) -- (prop.west);
  
  \draw[newtip,<-] ($(egocircle.south)+(0,0)$) -- ($(egocircle.south)+(0,-3em)$) -- 
  node[midway,anchor=south,xshift=0em,yshift=1em]{Iterate N Steps}($(prop.south) + (0,-3em)$) -- (prop.south);
  
  \draw[newtip] ($(cfs.north)+(0,1.6em)$) -- (cfs.north);
  \draw[newtip] (cfs.south) -- (score.north);

  \draw[newtip] (score.south) -- node [midway,left,xshift=-3pt,yshift=-0.5em]{Recent Best Trajectory}(controller.north);
  
  \draw[newtip,<-] ($(traj.south)+(0em,-4.6em)$) -- (traj.south|-score.east) |-
  node[midway,left,anchor=east,xshift=-5pt,yshift=19pt]{Replan}(score.east);

  \draw[newtip] ($(goal.east)+(3pt,0em)$) -- ($(goal.east)+(5.5em,0em)$);
  
  \draw[newtip] (kalman.south) -- (uncert.north);
  \draw[newtip] (uncert.east) -- (cfs.west);
  \draw[newtip] ($(uncert.north)+(2.5em,0)$) -- ($(uncert.north)+(2.5em,4em)$) -- ($(uncert.north)+(7.5em,4em)$);
%
%

  
  \node[dottedBlock,blueish,anchor=north west,minimum width=1em,text width=1em] (legend) at ($(laserscan.south west) + (1.5, -5.5)$) {};
  \node[anchor=west] (legend_name) at ($(legend.east) + (0.2, 0)$) {Contributions};
\end{tikzpicture}}
  \caption{Comparison between potential gap (PGap) in \cite{PGap} (left) and our proposed Hierarchical DAGap (H-DAGap) (right). The goal is given from a globally scaled problem, DAGap operates on locally scaled problems using paired agents, CFS enlarges the problem to all sensed agents, and SSA operates at every step with the unsafe agents.}
  \label{fig:pipeline}
  \vspace{-1em}
\end{figure*}
\begin{figure}
  \centering
  \hspace{-3em}
  \scalebox{0.8}{\usetikzlibrary{backgrounds}

\tikzstyle{block} = [draw, rectangle, text centered, thick,
                     minimum height=2em, minimum width=10em, inner sep=4pt]
\tikzstyle{reddish} = [draw=red,fill=white!95!red]
\tikzstyle{greenish} = [draw=green!40!gray,fill=white!95!green]
\tikzstyle{blueish} = [draw=blue,fill=white!95!blue]
\tikzstyle{newtip} = [->, very thick]

\begin{tikzpicture}[auto, inner sep=0pt, outer sep=0pt, >=latex]

  \node[centered] (kalman) {Kalman Filter};
  
  \node[centered,anchor=south,align=center] (uncert) at ($(kalman.north)+(0,3em)$) {Uncertainty \\ Analysis};

  \node[centered,anchor=west] (gap) at ($(uncert.east)+(5.5em,1.5em)$) {DAGap};
  
  \node[centered,anchor=west] (cfs) at ($(uncert.east)+(5.5em,-1.5em)$) {CFS};
  
  \node[centered,anchor=west] (ssa) at ($(kalman.east)+(9em,0em)$) {SSA};
  
  \begin{scope}[on background layer]
  \node[block,greenish,anchor=west,minimum height=2.5em, minimum width=19em] (control) at ($(kalman.west)+(-1.5em,0)$) {};
  
  \node[block,blueish,anchor=south west,minimum height=5em, minimum width=15em] (plan) at ($(uncert.south west)+(-1em,-1.5em)$) {};
  \end{scope}
  
  \draw[newtip] ($(kalman.west)+(-6.5em,0)$) -- ($(kalman.west)+(-0.1em,0)$)
  node[midway,anchor=south,xshift=-1em,yshift=0.5em]{Laserscan};
  
  \draw[newtip] (control.east) -- ($(control.east)+(5em,0)$)
  node[midway,anchor=south,xshift=0em,yshift=0.5em]{Control};
  
  \draw[newtip] ($(kalman.east)+(0.1em,0em)$) -- ($(ssa.west)+(-0.1em,0em)$);
  
  \draw[newtip] ($(kalman.north)+(0em,0.1em)$) -- ($(uncert.south)+(0,-0.1em)$);
  
  \draw[newtip] ($(uncert.east)+(0.1em,0em)$) -- ($(uncert.east)+(3.5em,0em)$)
  node[midway,anchor=north,xshift=0em,yshift=-0.5em]{$d_{safe}$} -- ($(cfs.west)+(-2em,0em)$) -- ($(cfs.west)+(-0.1em,0)$);
  
  \draw[newtip] ($(uncert.east)+(3.5em,0em)$) -- ($(gap.west)+(-2em,0em)$) -- ($(gap.west)+(-0.1em,0)$);
  
  \draw[newtip] ($(plan.east)+(0em,0em)$) -- (plan.east-|ssa.north) --
  node[midway,anchor=west,xshift=0.2em,yshift=0em]{Trajectory}($(ssa.north)+(0,0.1em)$);
  
  \node[centered,anchor=north] at ($(control.south)+(-0.8em,-0.8em)$) {Controller Thread};
  \node[centered,anchor=south] at ($(plan.north)+(0em,0.8em)$) {Planner Thread};
\end{tikzpicture}}
  \caption{Parallel computation architecture. The Kalman filter and SSA
  controller are run in one thread while the high-level planner is
  executed in another thread.\label{fig:parallel}}
\end{figure}
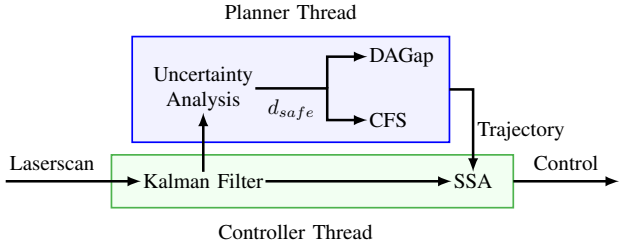

We first give an overview of PGap pipeline and our solution Hierarchical
DAGap (H-DAGap) to contrast the methods, see \cref{fig:pipeline}. The
PGap pipeline detects gaps, synthesizes trajectories for each gap by
following the gradient field, and picks the one with best score. PGap
navigation is designed find free space navigation affordances in static
environments, while DAGap extends the search for feasible trajectories
to spatio-temporal space. Moreover, a computational efficient trajectory
optimization method CFS and uncertainty analysis are exploited in the
planner to improve the safety and robustness of trajectories. 
High-level planning happens in separate thread with agent state
estimation and low-level safe control executed in the main thread, see
\cref{fig:parallel}.  The following subsections cover the details of
H-DAGap. 

\subsection{Dynamic Agent Gap Analysis}
\subsubsection{Inflated Agent Gap Detection}
\label{sec:agent gap detection}
When a new laser scan-like measurement $\mathcal{L}$ (360$^\circ$)--consisting of $n$ measurements of dynamic agents within the maximum sensing range $d_{\max}$--is available, we first pass the measurement into a Kalman filter module to estimate the agents' positions and velocities. Then we use $\mathcal{L}$ in our dynamic agent gap (DAGap) analysis module,  containing two components: inflated agent gap detection which leads to a set of gap $\rgpSet$ and dynamic gap analysis which synthesizes a trajectory for each gap $\gap \in \rgpSet$. 

Inflated agent gap detection inflates the range measurement $\mathcal{L}$ of agents by expanding the radii of the agents (to $\rIns$) to allow the robot to be treated as a point \cite{egoTEB}. To guarantee the distance between any points inside the gap region and the agents is larger than $\rIns$ when passing through the gap, we calculate the tangent points from the robot to the inflated circle as gap endpoints, see \cref{fig:inflated egocircle}. Each agent's inflation radius is adjusted based on its Kalman filter estimation covariance to ensure safety with higher probability (refer to \cref{sec:uncertainty}). Let $i$ be a circular index of the inflated measurement $\mathcal{L}^{inf}$, we perform a clockwise pass through $\mathcal{L}^{inf}$ and look for candidate gaps satisfying the following two requirements:
\begin{enumerate}
  \item Large range difference:
    $|\mathcal{L}^{inf}_l(i+1)-\mathcal{L}^{inf}_r(i)| > 2\rIns$
  \item Large angle difference:
    $|\theta_l(i+1)-\theta_r(i)| > \theta_{thre}$, 
\end{enumerate}
$l$, $r$ means the left or right tangent point, $\theta(i)$ is the scan
angle associated to index $i$ and $\theta_{thre}$ is a user defined
parameter. The candidate gap pairing uses the right tangent point of
previous agent and left tangent point of next agent. When there is a
wide open gap between two agents, agent1 and agent2 in \cref{fig:dgap1},
split it into multiple passable gaps by placing static virtual agents
at a user defined interval. If one agent or no agents are detected,
use the straight-line local planner targeting the goal.

\subsubsection{Dynamic Gap Analysis}

\begin{figure*}[!htb]
\minipage{0.65\textwidth}
  \begin{subfigure}{0.48\textwidth}
     \includegraphics[width=\textwidth,clip=yes,trim=120pt 50pt 120pt 70pt]{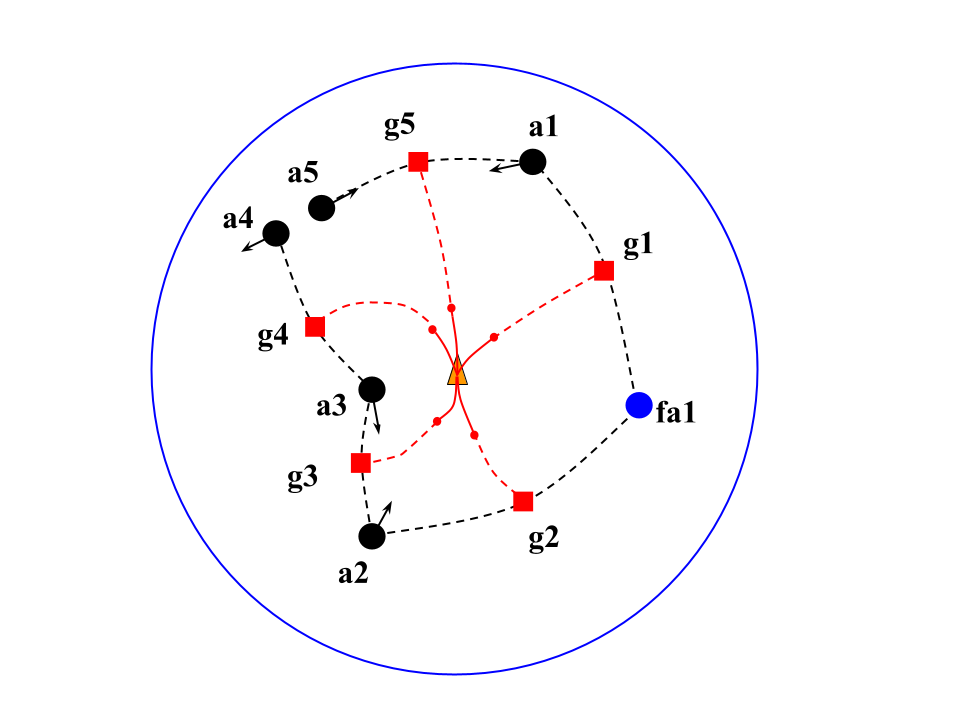}
     {\caption{DAGap synthesizes multiple trajectories (red, dashed lines) based on detected gaps.} \label{fig:dgap1}}
    \end{subfigure}
    \hfill
    \begin{subfigure}{0.48\textwidth}
     \includegraphics[width=\textwidth,clip=yes,trim=120pt 50pt 120pt 70pt]
       {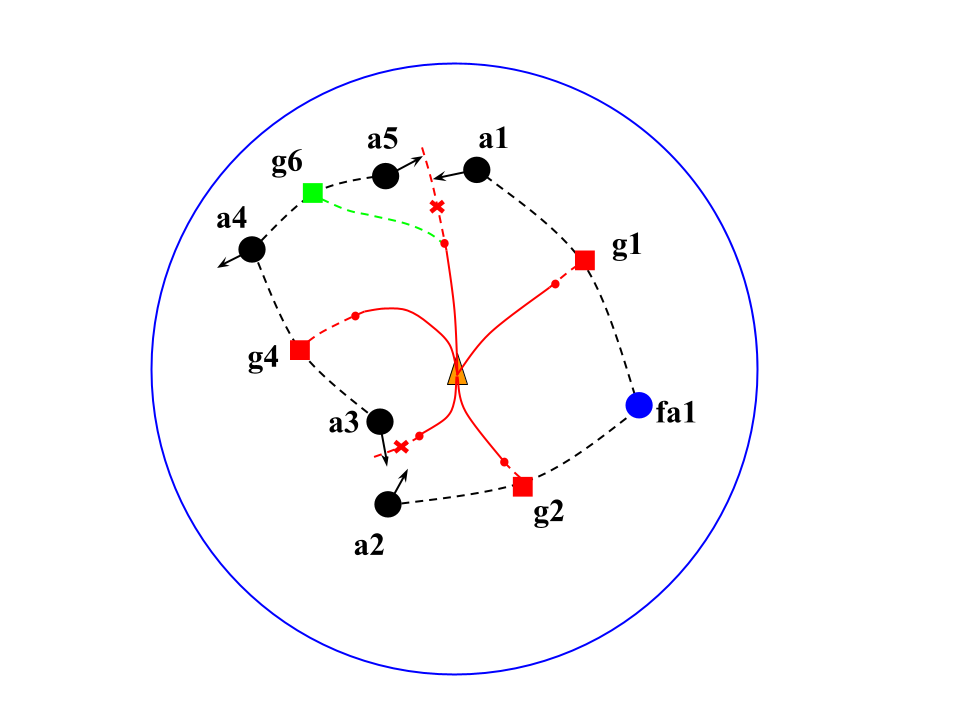}
     {\caption{DAGap trajectory updates for newly-opened (green) and closed gaps (red x).} \label{fig:dgap2}}
  \end{subfigure}
  \begin{subfigure}{\textwidth}
        \centering
        \includegraphics[width=0.70\textwidth]{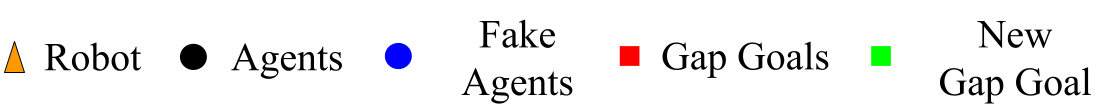}
        \label{fig:legend}
  \end{subfigure}
  \caption{DAGap synthesizes and updates trajectories.}
\endminipage\hfill
\minipage{0.32\textwidth}%
  {\adjustbox{trim=0pt 25pt 0pt 10pt}{\includegraphics[width=\linewidth]{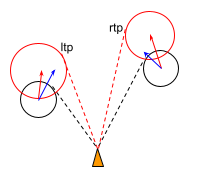}}}
  \caption{Inflated agent gap detection. Black and red circles are the initial and expanded inflated circles. Red and blue arrows are the estimated and true agent's velocity. Ltp and rtp mean left and right tangent point, respectively.}\label{fig:inflated egocircle}
\endminipage
\vspace{-1em}
\end{figure*}

Two problems need to be resolved. First, how to synthesize the trajectory to accommodate to the spatio-temporal dynamics of an open gap. For this, we predict the agents' positions based on the Kalman filter, construct a predicted scan $\mathcal{L}$, and recover predicted gaps and gap regions for $N$ steps into the future. Each gap defines a local gap goal. When all gap goals are connected to the robot, they define a star-like graph. $N$ repeated single-step iterations following the PGap gradient field \cite{PGap} using the predicted gaps synthesizes a set of candidate trajectories. This reactive approach is computationally fast. 

Predicted future gaps states may lead to a new gap or have a previously-open gap close. These birth/death events need to be resolved.  If a new gap is detected due to agents moving away from each other, like agent4 and agent5 in \cref{fig:dgap2}, create a trajectory for this newly-open gap at \cref{algo:dgap} line 9-10. From all existing trajectories, pick the one closest to the new gap as the path to expand from. This new trajectory will split from the old trajectory towards the new gap. In \cref{fig:dgap2}, the green trajectory to gap6 is copied from the trajectory to a previously-open but now closed gap5. The previously-open but now closed gap will be labelled as closed and trajectory updating will be halted. After planning finishes, return all trajectories targeting open gaps.

\begin{algorithm}[t]
	\caption{DAGap Trajectory Synthesis} 
	\label{algo:dgap}
	\begin{algorithmic}[1]
	\State $\trajSet \leftarrow TrajectorySet$
	\State $\rgpSet \leftarrow GapSet$
	\Function{DAGap}{$\mathcal{L}^{inf}$} 
		\For {$\texttt{horizon}=1,2,\ldots,N$}
            \State $\rgpSet \leftarrow$ gapDection($\mathcal{L}^{inf}$)
            \If{$\texttt{horizon} == 1$} 
		        \State initialize $\traj$ for $\gap \in \rgpSet$ and add into $\trajSet$
		    \EndIf
			\For {$\gap \in \rgpSet$}
			    \If{$\traj$ of $\gap$ doesn't exist in $\trajSet$} 
			        \State initialize $\traj$ for $\gap$ and add into $\trajSet$
			    \EndIf
                \State $\traj$ one step forward towards $\gap$ using PFM
			\EndFor
			\State update the agents' $\mathcal{L}^{inf}$ with kalman filter
		\EndFor \\
	    \Return $\trajSet$
	\EndFunction
	\end{algorithmic} 
\end{algorithm}

\subsection{Trajectory optimization and scoring}
A set of trajectories $\trajSet$ is generated from the \cref{algo:dgap}.  Each trajectory considers the specific pairwise agents forming its gap, which is acceptable in static environments. Dynamic agents can drive towards the robot outside the gap region. It's risky if the robot doesn't plan in advance, especially when agents move fast. Thus the second problem is to ensure safety constraint satisfaction for all agents over the planning horizon. The CFS optimizer, which can efficiently find optimal solutions that are strictly safe, is adopted to further modify these reference trajectories $\trajSet$. Compared to other optimization methods like sequential quadratic programming (SQP), CFS exploits problem geometry to improve computational efficiency while solving the optimization, which is critical for online planning \cite{CFS}.

For each trajectory, robot with initial pose $\bm x_0$ at time $t$ is suppose to reach a local goal position $\bm x_{lgoal}$ at time $t + T$ . Let $T = N \Delta t$ , where $N$ is the planning horizon and $\Delta t$ is the discrete time interval. The robot trajectory is denoted as $\bm s = [\bm x^{[0]} ; . . . ; \bm x^{[i]}; . . . ;\bm x^{[N-1]}]$, where $\bm x^{[0]} = \bm x_0$ , and $\bm x^{[N-1]} = \bm x_{goal}$. The trajectory of agent $j$ is denoted as $\bm s_O^j = [\bm o^{[0]}_j; . . . ;\bm o^{[i]}_j; . . . ;\bm o^{[N-1]}_j]$, where $j \in \{1,2,...,M\}$ and $M$ means the number of agents. Therefore, the robot should plan the trajectory that can reach the local goal while keeping the safety distance $\rIns$ from all agents for every time step. Mathematically, the discretized optimization problem is formulated as:
\begin{subequations}\label{eq:discretize cfs problem}
\begin{align}
\min_{\boldsymbol{s}} ~& \|\bm s - \bm s_r \|_{Q_{r}}^2 + \|\bm s \|_{Q_{s}}^2,\label{eq: discretize cost}\\
s.t. ~& D(\bm x^{[i]}, \bm o^{[i]}_j) \geq \rIns, \forall i, \forall j, \label{eq: discretize safety constraint}\\
& \bm x^{[1]} = \bm x_0, \bm x^{[M]} = \bm x_{lgoal}
\end{align}
\end{subequations}
where $\|\bm s - \bm s_r \|_{Q_{r}}^2$ penalizes the deviation from the new trajectory to the reference trajectory, and $\|\bm s \|_{Q_{s}}^2$ penalizes the properties of the new trajectory itself which ensures low velocity and acceleration magnitude. Constraint $D(\bm x^{[i]}, \bm o^{[i]}_j) \geq \rIns$ requires that the robot should keep safe at each planning step, where $D(.\,,.)$ computes the Euclidean distance between two points. The trajectory scoring \cref{eq:score} combines global target goal efficiency and optimization cost. The trajectory with highest score will be selected to follow:
\begin{equation}
\label{eq:score}
    J(\traj) = D(target, \bm x^{[M]}) - \|\bm s - \bm s_r \|_{Q_{r}}^2 - \|\bm s \|_{Q_{s}}^2 \\
\end{equation}

\subsection{Uncertainty Analysis and Replanning}
\label{sec:uncertainty}
In the above process, the safety distance is a fixed value $\rIns$ that will work with perfect prediction about agents' trajectories. In uncertain world, however, estimated agent positions and velocities from the Kalman filter have errors that will propagate as planning horizon increases. Using a fixed safety distance $\rIns$ may lead to collision.  To mitigate this problem, enlarge the safety distance according to the estimated high-confidence bound of prediction errors.  Assume the agent has constant velocity. Denote the current estimated agent state as $\bm z^{[0]} = \left[ \begin{matrix} \bm o^{[0]};\bm v \end{matrix}\right]$, including position $\bm o^{[0]}$ and velocity $\bm{v}$, and its estimated covariance matrix as $\bm{\Sigma_0}$. The ground truth state is labeled as $\bm{z^{[0]}}_{*}$. The future agent state predicted at step $i$ is $\bm z^{[i]} = F^i \bm z^{[0]}$, where $F$ represents the transfer matrix. Suppose $\bm z^{[0]} \sim \mathcal{N}(\bm z^{[0]}_{*},\, \bm \Sigma_0)$, the prediction for $\bm z^{[i]}$ should follow the distribution: $\bm z^{[i]} \sim \mathcal{N}(\bm z^{[i]}_{*},\,\bm{\Sigma_i})$, where $\bm{\Sigma_i}$ is the covariance matrix propagated forward for $i$ steps $\bm{\Sigma_i} = (F^T)^i \bm{\Sigma_0} F^i$. Pick out the submatrix of $\bm{\Sigma_i}$ corresponding to the position $\bm o^{[i]}$ and denote the position covariance matrix as $\bm{\Sigma_{i,o}}$. Then the error $\Delta \bm o^{[i]} = \bm o^{[0]} - \bm o^{[0]}_{*}$ should follow the chi-square distribution ${\chi}^2_{N}$ as \cref{eq:chi-square}, where $N$ is the dimension of $\bm{o^{[i]}}$, 
\begin{equation} \label{eq:chi-square}
    (\bm{\Delta o^{[i]})}^T \, \bm{\Sigma_{i,o}}^{-1} \,\bm{\Delta o^{[i]}} \sim {\chi}^2_{N}
\end{equation}
with the probability support on confidence bound value $k_{\epsilon}$: 
\begin{equation} \label{eq:confidence}
    P((\bm{\Delta o^{[i]})}^T \, \bm{\Sigma_{i,o}}^{-1} \,\bm{\Delta o^{[i]}} \leq k_{\epsilon}) > 1 - \epsilon
\end{equation}
based on Lemma 4 in \cite{cheng2020safe}. The following bound on the error $\bm{\Delta o^{[i]}}$ holds with probability $1 -\epsilon$:
\begin{equation}
\label{eq:discretize cfs problem}
-\sqrt{k_{\epsilon} \lambda_n} \leq \bm{v_n}^T \bm{\Delta o^{[i]}} \leq  \sqrt{k_{\epsilon} \lambda_n}, \forall n\\
\end{equation}
where $\{\lambda_n\}’s$ and $\{\bm{v_n}\}’s$ are the eigenvalues and eigenvectors of $\bm{\Sigma_{i,o}}$, $n \in \{1,2,...,N\}$. Since the $\{\bm{v_n}\}’s$ are perpendicular bases, $\bm{\Delta o^{[i]}}$ can be represented as $\sum_n a_n \bm{v_n}$, where $a_n$ is the coefficient, 
\begin{subequations}
\begin{align}
&\bm{v_n}^T \bm{\Delta o^{[i]}} = \bm{v_n}^T \sum_k a_k \bm{v_k} = a_n \|\bm{v_n}\|\\
&-\frac{\sqrt{k_{\epsilon} \lambda_n}}{ \|\bm{v_n}\|} \leq a_n \leq  \frac{\sqrt{k_{\epsilon} \lambda_n}}{ \|\bm{v_n}\|}\\
&\|\bm{\Delta o^{[i]}}\| = \|\sum_n a_n \bm{v_n} \| \leq \sum_n\sqrt{k_{\epsilon} \lambda_n}
\label{eq:bound}
\end{align}
\end{subequations}

\begin{theorem}
\label{theorem:uncertainty}
Using the bounds in \cref{eq:bound}, increasing the safety distance by $\sum_n \sqrt{k_{\epsilon} \lambda_n}$ guarantees safety with probability at least $1-\epsilon$. 
\end{theorem}

\newcommand\qedsymbol{$\blacksquare$}

\begin{proof}
Since the prediction of the ground truth agent's position $\bm o^{[i]}_{*}$ is unbounded, we ensure a probability safety constraint between $\bm x^{[i]}$ and $\bm o^{[i]}_{*}$ for $\forall i$, 
\begin{equation}
    P(\|\bm x^{[i]} - \bm o^{[i]}_{*}\| \geq \rIns) < 1 - \epsilon
\end{equation}
At worst case, $\|\bm x^{[i]} - \bm o^{[i]}_{*}\| = \|\bm x^{[i]} - \bm
o^{[i]} + \bm o^{[i]} - \bm o^{[i]}_{*}\| = | \|\bm x^{[i]} - \bm
o^{[i]}\| - \|\bm o^{[i]} - \bm o^{[i]}_{*}\| |$. From \cref{eq:bound},
the bound on the uncertainty $\|\bm o^{[i]} - \bm o^{[i]}_{*}\|$ holds
with probability $1 -\epsilon$. Expanding the safety distance $d_{safe}$
between the robot and the estimated agent position $\|\bm x^{[i]} - \bm
o^{[i]}\|$ to $\rIns + r$, where $r = \sum_n\sqrt{k_{\epsilon}
\lambda_n}$, guarantees safety with probability at least $1 -\epsilon$.
\end{proof}

Based \cref{theorem:uncertainty}, estimate the robust safety distance
$d_{safe}=\rIns + r$ for every agent and replace the fixed $\rIns$ used
in DAGap and CFS (see \cref{fig:parallel}).  As a detection threshold,
larger values of $d_{safe}$ increase the likelihood of false negatives
(e.g., rejection of passable paths), which limits the planning space and
leads to more conservative behavior.  To avoid this problem, we set an
upper bound of the $d_{safe}$ and record the first time step $k$ this
upper bound is reached. Replanning is evoked after executing $k$ steps
due to low safety likelihood, see \cref{algo:H-DAGap} line 18.

\begin{algorithm}
	\caption{H-DAGap Algorithm} 
	\label{algo:H-DAGap}
	\begin{algorithmic}[1]
	\State $\uncertainAnalysis \leftarrow uncertaintyAnalyzor$
	\State $\safeDistanceSeq \leftarrow$ robust safe distances $d_{safe}^i$, $i \in \{1,2...N\}$
	\State $\texttt{k} \leftarrow$ replan step

	\Function{Traj Optimization}{$\trajSet$, $\uncertainAnalysis$} 
	    \For {$\traj \in \trajSet$}
            \State compute $\safeDistanceSeq, \texttt{k}$ with $\uncertainAnalysis$
            \State CFS optimizes and scores $\traj$ using $\safeDistanceSeq$
        \EndFor
        \State \Return $\traj$ with highest score and its $\texttt{k}$
    \EndFunction
    \Statex
    \While {not arrive goal area}
        \State $\trajSet$  $\leftarrow$ \Call{DAGap}{$\mathcal{L}^{inf}$}
        \State $\traj, \texttt{k}$  $\leftarrow$ \Call{Traj Optimization}{$\trajSet$, $\uncertainAnalysis$}
        \State $\texttt{stepCount}$ = 0
        \For {next waypoint $\texttt{p}$ in $\traj$}
            \State safe control $\texttt{u}$ $\leftarrow$ SSA($\texttt{p}$)
            \State execute $\texttt{u}$ and $\texttt{stepCount} \mathrel{+}= 1$
            \State break for replan if $\texttt{stepCount} \geq \texttt{k}$
        \EndFor
    \EndWhile
	\end{algorithmic} 
\end{algorithm}

\subsection{Safe Controller} 
Even though DAGap and CFS are efficient planning methods, their longer horizon mean that they are more computationally expensive compared to one-step reactive safe control methods. Moreover, challenging crowded environments mean that CFS may not have converged within a few iterations. Due to the real-time planning requirement, we don't run CFS for all synthesized trajectories until they converge. Instead we select the top two trajectories based on the efficiency score $D(target, \bm x^{[M]})$ as we notice the optimization cost doesn't change the rank ordering of top two candidates in most cases, and run CFS for only one iteration. Moreover, we adopt fast SSA in the low-level controller layer to further compensate the long planning time, and monitor the control at every step. Compared to other reactive algorithms like CBF which enforces constraints everywhere, SSA achieves better safety-efficiency trade-off in complex environment \cite{RL_SSA}. 

The key of SSA is to define a valid safety index $\phi$ such that 1) there always exists a feasible control input in control space that satisfies $\dot\phi\leq -\eta\phi$ when $\phi \geq 0$ and 2) any control sequences that satisfy $\dot\phi\leq -\eta\phi$ when $\phi \geq 0$ ensures forward invariance and asymptotic convergence to the safe set $X_S$, $\eta$ is a positive constant that adjusts the convergence rate. In our problem, $X_S = \left\{x | \phi_0(x) \leq 0 \right\}$, where $\phi_0$ is defined as $d_{min}^2 - d^2$, $d_{min}$ is the user defined minimal distance and $d$ is the distance from the robot to the agent. Since the robot we adopt in testing is a second-order system, we add higher order term of $\phi_0$ to ensure that relative degree one from safety index $\phi$ to the control input, and $\phi$ is defined as follows:
\begin{equation}\label{eq: safety index}
\phi = d_{min}^2 - d^2 - k\cdot \dot{d}.
\end{equation}
where $\dot{d}$ is the relative velocity from the robot to the agent and
$k$ is a constant factor. As proved in \cite{SSA}\cite{SSA_Zhao}, the
safety index $\phi$ will ensure forward invariance of the set $\phi \leq
0 \cap \phi_0 \leq 0$ and global attraction to that set. With safety index $\phi$, project the reference control $u^r$ to the set of safe controls that satisfy $\dot{\phi} \leq -\eta\,\phi$ when $\phi \geq 0$, and $\dot{\phi}$ is expressed as
\begin{equation}
    \dot \phi = \frac{\partial \phi}{\partial x}\,f + \frac{\partial \phi}{\partial x}\,g\; u = 
    L_{f}\phi+L_{g}\phi\; u.
\end{equation}
Compute $\phi_{j}$ for every agent and add the safety constraint whenever $\phi_{j}$ is positive.  SSA solves the following one-step optimization problem, with safety and dynamics constraints, through quadratic programming (QP) when triggered: 
\begin{subequations}
\begin{align}
    \underset{u\in U}{\min}\,&||u-u^r||^2 
    = \underset{u\in U}{\min}\,u^\mathrm{T}\left[ \begin{matrix}
	1&0\\
	0&1
    \end{matrix}\right]u - 2u^\mathrm{T}\left[ \begin{matrix}
	1&0\\
	0&1
    \end{matrix}\right]u^r \label{eq:ssa}\\
    &s.t. L_{f}\phi_{j}+L_{g}\phi_{j}~u \leq -\eta\,\phi_{j}, j \in \{1,2...M\}. \label{eq:safe constraints}
\end{align}
\end{subequations}

\section{Experiments}
This section covers the experiments, results, and comparison of H-DAGap with ARENA \cite{ARENA} and DRRT-ProbLP \cite{DRRT-ProbLP}. 

\subsection{Benchmark Results}
\begin{figure}
  \centering
  \vspace{0.5em}
  \includegraphics[width=0.38\textwidth,height=6.5cm]{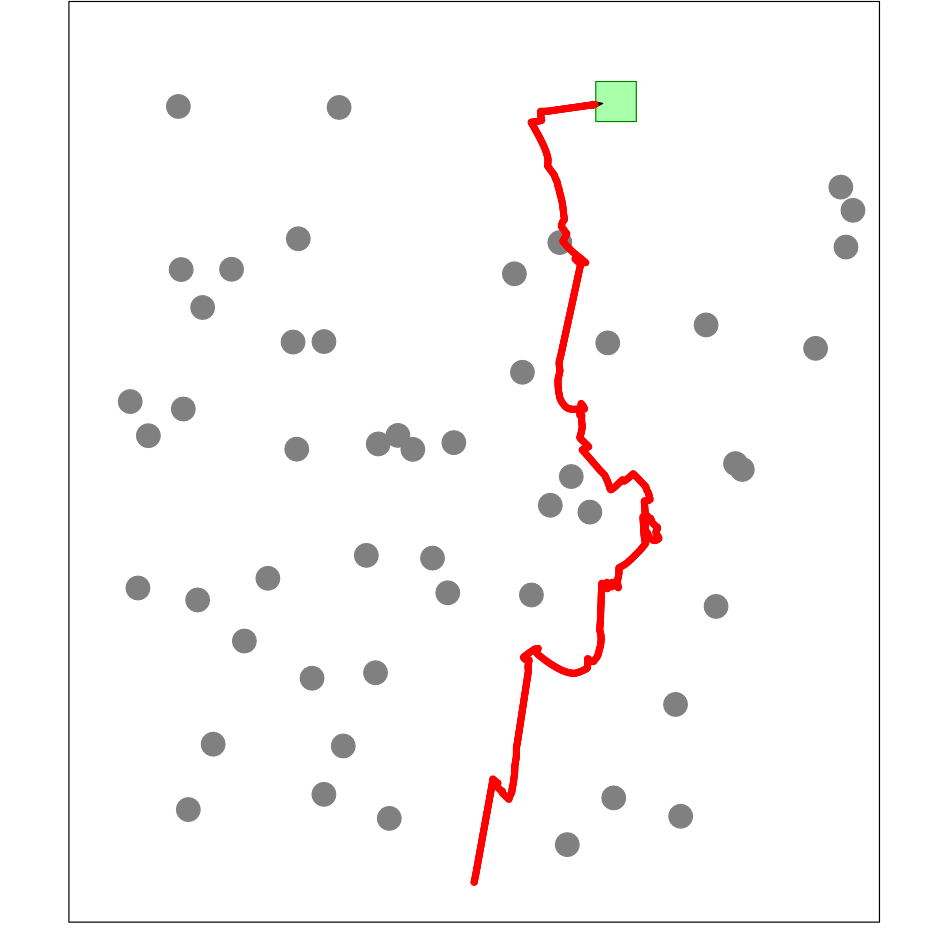}
  \caption{Testing environment with 50 dynamic agents. The green square is the target goal, the gray circles are agents' positions at last time step and the red path is the entire robot trajectory from the start to the goal.}
  \label{fig:environment}
\vspace{-5pt}
\end{figure}

Experiments are conducted in a $2 \times 2$ empty world, see
\cref{fig:environment}, with 20 and 50 dynamic agents in two test
scenarios. ARENA nd DRRT-ProbLP are tested in 20 agents environments
\cite{ARENA}\cite{DRRT-ProbLP}, and we test our model in the same
condition for comparison and also increase the agents' number to 50 to
test their capability in more challenging scenario. The radius of agents
is 0.05 and the velocity is sampled from a uniform distribution $\left[
5e^{-3}, 2e^{-2}\right]$. Agents can move randomly in any direction in
the map and collisions between them are not considered. The robot adopts
the second order unicycle model and with speed range $\left[ 0,
2e^{-2}\right]$. The overall scenario settings used in ours and two
baseline papers are similar, including the relatively agent size to
world size and the relatively agent velocity to robot velocity. But we
don't assume the perfect lidar measurement is available, instead, the
360$^\circ$ field of view (FoV) measurement has errors following the
Gaussian distribution $\mathcal{N}(0,\,0.01^{2})$. We use Kalman filter
to track the position and velocity of each agent inside the $0.2$
sensing range. Compare to the environment used in previous safe learning
work \cite{RL_SSA}, we enlarge the agent size, increase its top speed,
and add measurement uncertainty, which makes the task more challenging.
H-DAGap is run in Python on Ubuntu 20.04 of 3.7 GHz using Intel Core i7.
The average computate times for DAGap, CFS and SSA are $0.0814s$,
$0.1120s$ and $6.854\mathrm{e}{-4}s$ respectively. We conduct 100 test
runs in each scenario and use collision rate and success rate as
evaluation metrics. A success trial means the robot reaches the goal
within 3500 steps without any collision. The robot is allowed to
continue driving after collision in benchmark papers. We follow this
rule but don't observe multiple collisions in any H-DAGap trial.  

ARENA and DRRT-ProbLP are used for safe navigation in crowded dynamic environments and tested in environment with 20 dynamic agents in their papers. H-DAGap achieves $97\%$ success rate and only $3\%$ collision rate in the 50 dynamic agents environment (see \cref{tab:safety comparison}), which bests the benchmark implementations in the 20 agent environment. For DRRT-ProbLP planner, the DRRT module only considers static obstacles and the avoidance of dynamic agents is entirely handled by ProbLP. ProbLP samples the robot movement direction based on the distribution of target goal and agents, then synthesizes and scores the trajectories. However, there are usually many safe trajectories in open continuous space; Sampling-based methods can't enumerate all of them and inevitably become suboptimal. ARENA combines the traditional global planner A* and the Deep Reinforcement Learning based (DRL) local planner. Vanilla A* doesn't consider the dynamic agent and DRL doesn't consistently ensure safety constraint satisfaction during execution. On the other hand, H-DAGap considers and probabilistically guarantees safety in both layers and in multi-modules. 

\begin{table}[h!]
\vspace{0.5em}
\fontsize{6.5}{9}\selectfont
\centering
\caption {H-DAGap and benchmark algorithms outcomes in an empty world with 20 and 50 dynamic agents.}
\label{tab:safety comparison}
\begin{tabular}{@{}ccccc@{}}
\toprule
  & \multicolumn{2}{c}{20 agents} & \multicolumn{2}{c}{50 agents}\\
  Model & Success & Collision & Success & Collision\\ \midrule
{H-DAGap} & 100\% & 0\% & 97\% & 3\% \\\midrule
{DRRT-ProbLP}& N/A & 4\% & N/A & N/A \\\midrule
{ARENA} & $\leq$ 92.7\% \tablefootnote[4]{Note: The success criteria in ARENA is goal attainment with less than two collisions.} & 23.7\% & N/A & N/A \\
\bottomrule
\end{tabular}
\vspace{-10pt}
\end{table}

\subsection{Discussion}
This section looks into the contribution of different modules. Each experiment is repeated 100 times. Here, the test run will stop once the robot collides.

\textbf{DAGap trajectory synthesis result:} We compare the results of DAGap only and static gap detection (SGap), which synthesize trajectories for the gaps detected at current time step. CFS optimization and SSA modification are not applied. Compared to SGap, DAGap reduces the collision rate by $13\%$ and $7\%$ in 20 and 50 agent environments respectively, see \cref{tab:modular safety result}. The reason is that DAGap considers the spatio-temporal dynamics of open gaps and filters out the trajectories towards gradually closing gaps. SGap guarantees safe passage in static environments, however, the originally open gap may become closed and can lead to collision in dynamic environments. The improvement of incorporating spatio-temporally evolving information is greater for the easy scenario because DAGap is less affected by the agents outside the gap region due to the lower agents density.

\begin{table}[htbp]
\fontsize{7.5}{9}\selectfont
\centering
\caption {Experimental results of each module.}
\label{tab:modular safety result}
\begin{tabular}{@{}ccccccc@{}}
\toprule
& \multicolumn{2}{c}{20 agents} & \multicolumn{2}{c}{50 agents}\\ 
 Model & Collision & Success & Collision & Success\\ 
 \midrule
{SGap} & 49\% & 51\% & 76\% & 24\%\\
{DAGap} & 36\% & 64\%& 69\%& 31\%\\
{DAGap+CFS} & 8\%& 92\%& 29\%& 71\% \\
{DAGap+CFS+SSA} & 0\%& 100\%& 3\%& 97\% \\
\bottomrule
\end{tabular}
\end{table}

\textbf{CFS trajectory optimization result:} The collision rate drops from $69\%$ to $29\%$ in the 50 agent scenario after CFS optimization and uncertainty-based safety distance adjustment. Compared to using CFS directly, DAGap provides good initial trajectories that can improve the safety of optimized trajectories. To better explain it, we need to define the feasibility of a trajectory: a feasible trajectory requires the distances between all neighbouring waypoints be smaller than a threshold value related to the robot's maximal velocity. An infeasible trajectory increases the risk of collision and is hard to be tracked by robot due to the large jump between waypoints. The feasibility rate of CFS optimized trajectory is around $87.4\%$ when using DAGap to generate initial reference trajectory that drives towards the affordance free space, but is only $66.9\%$ without DAGap. 

\textbf{SSA safe controller result:} By replacing the feedback controller with the SSA safe controller, the collision rate drops to $3\%$. There are two main reasons behind: first of all, as we discussed above, CFS may generate dynamically infeasible trajectory due to the limited number of optimization iterations we can run in real-time and the challenging crowded dynamic environment. Tracking infeasible trajectory can cause collision. SSA modifies these unreasonable tracking controls online. Secondly, DAGap and CFS are long-term planners considering N steps safety. But the predicted error of trajectories of agents will compound as time goes, making the planned trajectory risky in the long-term future even we expand the safety distance. On the other hand, because of its fast computation property, SSA always uses the latest information to calculate the one-step safe control and to reduce collision caused by uncertainty. 

\begin{figure}
    \vspace{0.5em}
     \centering
     \begin{subfigure}{0.23\textwidth}
         \centering
         \includegraphics[width=\textwidth]{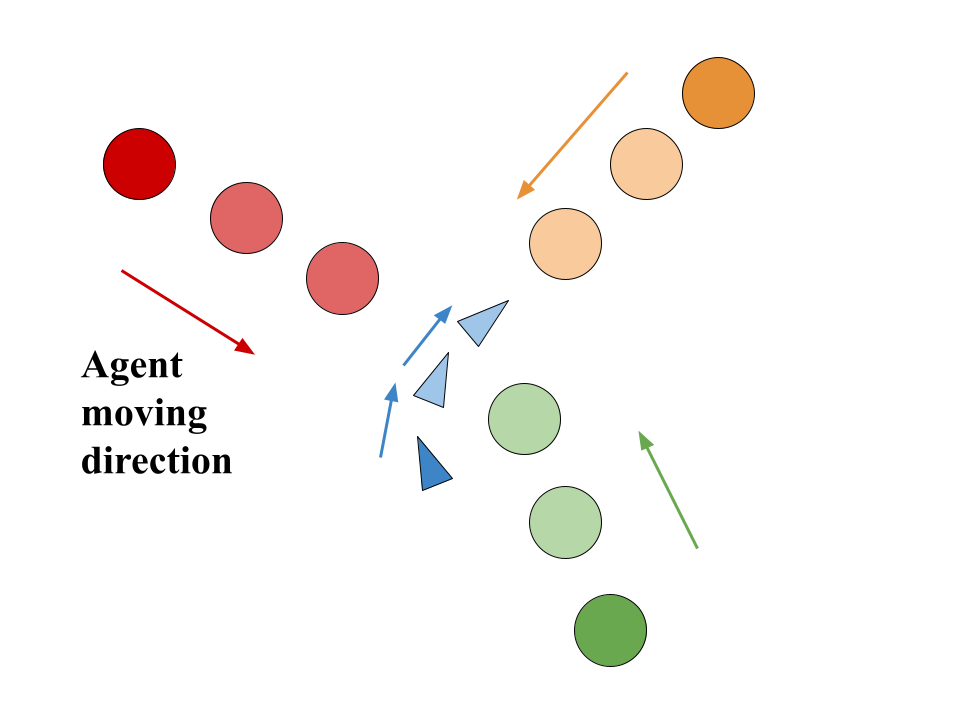}
         \caption{Trapping collision}
         \label{fig:trapping}
     \end{subfigure}
     \begin{subfigure}{0.23\textwidth}
         \centering
         \includegraphics[width=\textwidth]{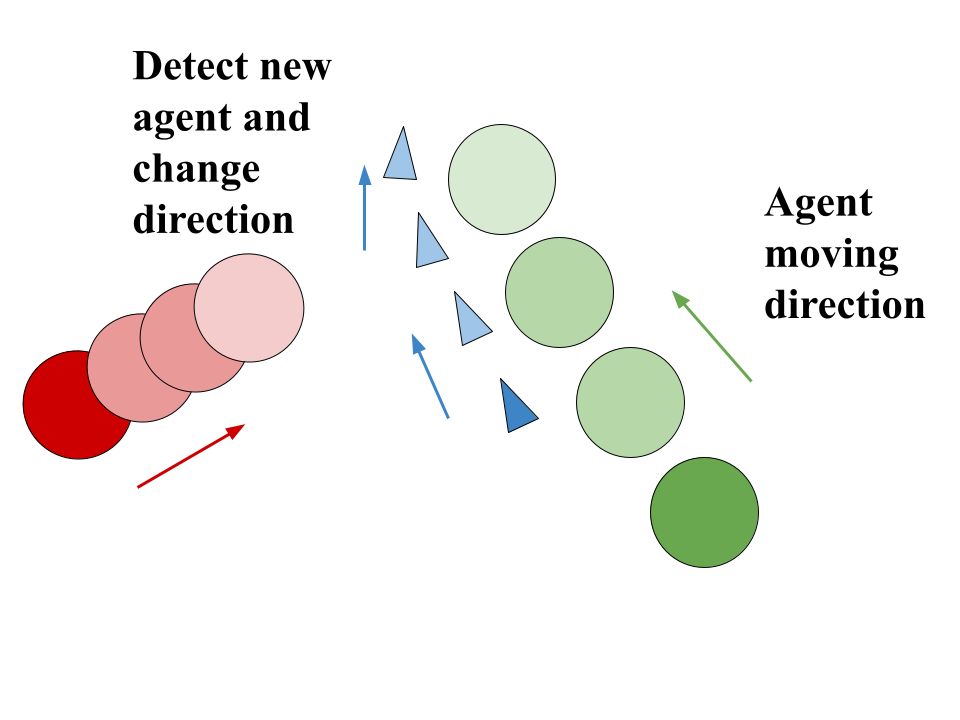}
         \caption{Overtaking collision}
         \label{fig:catchup}
     \end{subfigure}
     \caption{Trapping collision and overtaking collision. The robot is the blue triangle and the agents are colored circles. All are a darker hue at their initial positions and lighten as time progresses.}
     \label{fig:three graphs}
     \vspace{-1em}
\end{figure}

\label{sec:discuss}
\textbf{Collision analysis:} Even applying all these techniques, there
is still a $3\%$ collision rate. The collisions are categorized into two
main cases: multi-agent traps and a fast, overtaking agent. Notice, the agents in our environments are artificial and the collisions between agents are allowed. In the first case, trapping occurs when a gap detected as passable is actually not
passable (i.e., a false positive) or when an existing or future gap is
not detected due to sensing radius limits. The robot ends up trapped by
multiple converging agents, see \cref{fig:trapping}, and SSA cannot find
a control to meet all safety constraints because the robot will get
closer to one of the agents no matter in which direction it drives. 
The ``best" control for SSA is to stay put.
The second case happens when a fast agent driving behind
the robot and in collision overtakes it.
Following the safest one-step control generated by SSA, moves the robot
in a direction aligning with the agent's velocity even if the
DAGap trajectory points in another direction. Alignment occurs because
SSA modifies the original control to the safest single-step one based on
the safety index. This escape-and-pursue situation usually continues for
several steps until the robot meets another agent and needs to take a
new control to avoid both. The fast agent behind will then catch
up such that the robot cannot bypass both within one or two steps due to its size and
speed.  From the perspective of SSA, no control exists to satisfy the
constraints in \cref{eq:safe constraints} once the new agent enters the
sensing radius. 
%
The situations can be avoided through high-level modifications. 
One high-level planner design change would be to check if there may be
future trapping situations, then specify an alternative detouring
(global) goal until the danger is resolved. Additionally better
coordination between the high-level planner layer and low-level SSA
layer can avoid conflicts related to the safety specifications. We leave
these to future improvements.

\section{Conclusion}
This work described and evaluated H-DAGap, a hierarchical navigation
solution containing a multi-phase planner and a low-level safe
controller. It is a solution strategy to the safe navigation problem in
crowded, dynamic and uncertain environments. Estimated high-confidence
error bounds are used in the planner to achieve provably high
probability safety to uncertainty. Conducted experimental benchmarking 
in simulation and analysis confirm the effectiveness of H-DAGap at
avoiding collisions and navigating to the goal. The H-DAGap implementation 
is available at https://github.com/hychen-naza/H-DAGap.  Improved
coordination between the high-level planning and low-level safety
control to improve collisions in low-probability robot-agent
configurations is left to future work.  Extension of H-DAGap to consider
the case of reduced field of view sensing is also left to future
work.

\bibliographystyle{IEEEtran}
\bibliography{sample.bib}
\end{document}